\documentclass{llncs} 
\usepackage{graphicx}
\begin{document}
\title{Fast Dictionary Matching for Content-based Image Retrieval}
\author{Patryk Najgebauer\inst{1}, Janusz Ryga{\l}\inst{1}, Tomasz Nowak\inst{1}, Jakub Romanowski\inst{1}, Leszek Rutkowski\inst{2}, Sviatoslav Voloshynovskiy\inst{3}, Rafa{\l} Scherer\inst{1}}

\institute{Institute of Computational Intelligence, Cz\c{e}stochowa University of Technology\\
al. Armii Krajowej 36, 42-200 Cz\c{e}stochowa, Poland (\texttt{http://iisi.pcz.pl})
\and
University of Social Sciences in {\L}\'od\'z\\
Sienkiewicza 9, 90-113 {\L}\'od\'z, Poland \\
\and
University of Geneva, Computer Science Department, \\
7 Route de Drize, Geneva, Switzerland\\
\email{\{patryk.najgebauer, janusz.rygal, tomasz.nowak, jakub.romanowski, leszek.rutkowski, rafal.scherer\}
@iisi.pcz.pl}
}

\maketitle
\begin{abstract}
This paper describes a method for searching for common sets of descriptors between collections of images. The presented method operates on local interest keypoints, which are generated using 
the SURF algorithm. The use of a dictionary of descriptors allowed achieving good performance of the content-based image retrieval. The method can be used to initially determine a set of similar pairs of keypoints between images. For this purpose, we use a certain level of tolerance between values of descriptors, as values of feature descriptors are almost never equal but similar between different images. 
After that, the method compares the structure of rotation and location of interest points in one image with the point structure in other images. Thus, we were able to find similar areas in images and determine the level of similarity between them, even when images contain different scenes.  
\keywords{content-based image retrieval, local interest points, image matching}
\end{abstract}
\section{Introduction}
Content-based image analysis is important part of many areas of science and engineering. It can be used for face recognition \cite{Akhtar:Rattani:Foresti:JAISCR:2014}, medical imaging  \cite{Bruzdzinski:Krzyzak:Fevens:Jelen:JAISCR:2014}\cite{Karimi:Krzyzak:JAISCR:2013}\cite{Makinana:Malumedzha:Nelwamondo:JAISCR:2014}, military science \cite{Wang:Japkowicz:Matwin:JAISCR:2014} and general purpose image analysis \cite{Chu:Krzyzak:JAISCR:2014}.
Image comparison based on their content is a complex process and still far from the excellence of the human vision. The main problem is a difference between human perception and that what can be analysed by computers. Humans focus on the remembering semantic description of an image without details as well as events, actions and objects represented by images. At the same time, they are not able to reconstruct exactly what they saw, as we remember overall image context. Humans recognize image objects by linking the situation presented on the image and the information about it learned previously.
Computers cannot  find simply a relationship between images and the semantic description of objects being the content of the image. On the other hand, they are very detailed in the analysis and  thanks to their precision they are able to do things such as fingerprint\cite{DrozdaSG13} or  signature \cite{zalas2013} recognition   much faster and better than humans.  

 	Algorithms generating local keypoints are often used for general purposes. Unfortunately they generate huge amount of information, which  has to be compared afterwards. This feature implicates large complexity of computation both while generating and comparing keypoints.    
 Using keypoints, we can focus on certain areas of the image and skip the rest.

Keypoints have also spatial relationships, which are useful in comparison of similar areas in images regarding to their rotation and shift in different images. We cannot simply search for each keypoint in an image for its equivalent in the other image as some keypoints could exist only in one image.   
   
\subsection{SURF Algorithm}

\begin{figure}
\centering
\includegraphics[width=6cm]{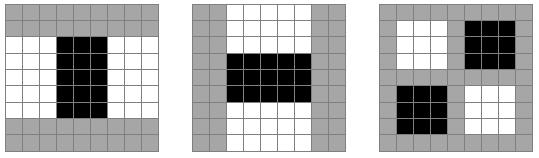}
\caption{SURF smallest 9x9 pixels blob detector}
\label{fig:Filters}
\end{figure}

\begin{figure}
\centering
\includegraphics[width=4cm]{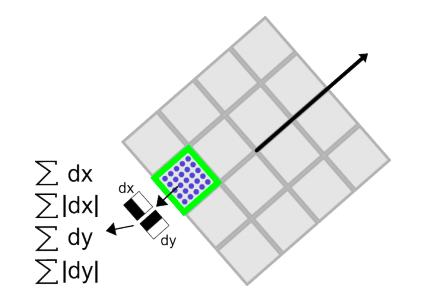}
\caption{Descriptor structure, 4x4 matrix of 4 values from subregion ($V_{sub}$)}
\label{fig:Surf1}
\end{figure}

One of the fastest algorithms nowadays for local interest point detection and description is  the SURF algorithm \cite{Bay:SURF}\cite{pena2012comparative}. One of the advantages of SURF over other algorithms is good performance, which is achieved by comparing areas in images and not single pixels. 
This approach has also a disadvantage: it is impossible to divide an image into circular areas.  The algorithm can function only with square areas, which can cause some inaccuracies in estimated values, when an image is for example rotated 45 degrees. 
The algorithm searches for areas, in which local values of second derivatives are the highest, i.e. for local extremes. It is a typical function of algorithms of the blob detection family (Fig. \ref{fig:Filters})\cite{Wang:BlobDetection}. It also estimates the size and the rotation of the keypoint, which is important for the exploration of dependencies between keypoints. 
The most important information, which is generated for each keypoint, is its descriptor. The descriptor allows to recognize a keypoint and represents local gradient around the keypoint. 
In the basic version of the SURF algorithm, the descriptor is built from 64 floating-point values. These values are grouped in 4 element chunks, which describe each of sixteen subregion of the descriptor: X-axis derivatives, Y-axis derivatives, modulus of the X-axis derivatives and modulus of the Y-axis derivatives 
\begin{equation}
\label{eq:IntImages}
V = \left(\sum dx , \sum dy, \sum|dx|, \sum|dy|\right) \;.
\end{equation}
Subregions create a 4x4 matrix that cover the keypoint localization in image (Fig. \ref{fig:Surf1}).

\section{Description of the Problem}

\begin{figure}
\centering
\includegraphics[width=4cm]{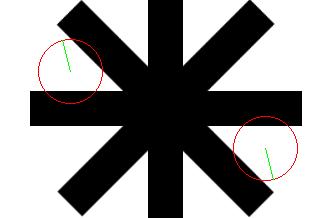}
\caption{An example of similar SURF  keypoints  with 0.47 value of difference between descriptor components}
\label{fig:Example1}
\end{figure}

As already mentioned, comparison of descriptors has to be done by  tresholding their difference, otherwise it will be almost impossible to match them to keypoints from a different image. For example, Table  \ref{Table1} presents distribution of differences in values of similar descriptors 
(Fig. \ref{fig:Example1}) with the sum of absolute differences (SAD, $L^1$ norm) equal to 0.4753. In this case, we consider the keypoints with SAD lower than 0.5 as similar.  

\begin{table}
\centering
\caption{Differences between $V_{sub}$ of two similar keypoints descriptor}
\begin{tabular}{     c     r     r     r     r     }
\hline{\smallskip} 
$V_{sub}x/y$ & 1 & 2 & 3 & 4\\
\noalign{\smallskip}
\hline
\noalign{\smallskip}
1 & 0.0000 & 0.0059 & 0.0031 & -0.0047 \\
2 & -0.0098 & 0.0144 & 0.0349 & 0.0159 \\
3 & -0.0495 & -0.0214 & -0.0159 & 0.0079 \\
4 & -0.0770 & -0.0062 & -0.0120 & -0.0173\\
\hline
\end{tabular}
\label{Table1}
\end{table}
Presented keypoints and their orientations are identical for humans, but according to the values of descriptors, they are different. 
After generation of keypoints in the process of comparison of two images, we have two sets of keypoints. The number of keypoints depends on the size of images and the amount of details. Often, for images larger than 1280x800 pixels, the number of keypoints exceeds 1000. The easiest and the most common approach of comparison of keypoints between images is to compare each keypoint with the rest, but when we deal with a large number of keypoints, the number of needed computations is very high. For example, 1000 of key points implicates 1 million of comparisons. To reduce the number of comparisons, keypoints should be ordered somehow and some of them should be passed over during the comparison process. 

Another challenge by the estimation of similar parts is the problem of  keypoints being lost during image transformations. The cause for this problem is different configuration of the same keypoints after the transformation. Usually images representing the same content contain only part of similar keypoints, another image can contain of course a different set of keypoints.  

\section{Method Description}
For better performance the proposed method uses a special, dictionary-based form of keypoint representation \cite{edelkamp2011heuristic}\cite{DBLP:NRGKS14}. Dictionary-based structure accelerates the comparison process by allowing to skip most of keypoint combinations.   

\subsection{Dictionary Creation}
\begin{figure}
\centering
\includegraphics[width=6cm]{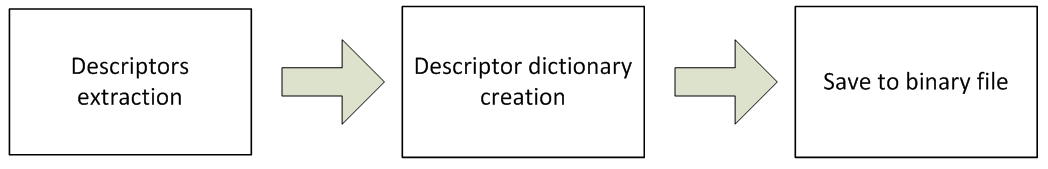}
\caption{Flowchart presenting the process of dictionary creation }
\label{fig:Shema1}
\end{figure}

Before performing matching images, the method prepares images by keypoint detection and generating the dictionary structure for each single image (see Fig. \ref{fig:Shema1}). 

Descriptor dictionary is created from 64 element vectors which are local interest point descriptors of an image. The method puts separate elements of the descriptor in the dictionary beginning from the first element. The dictionary is built in a similar way to the B-tree, where the first element of dictionary contains the list of first elements of descriptors.

\begin{figure}
\centering
\includegraphics[width=7cm]{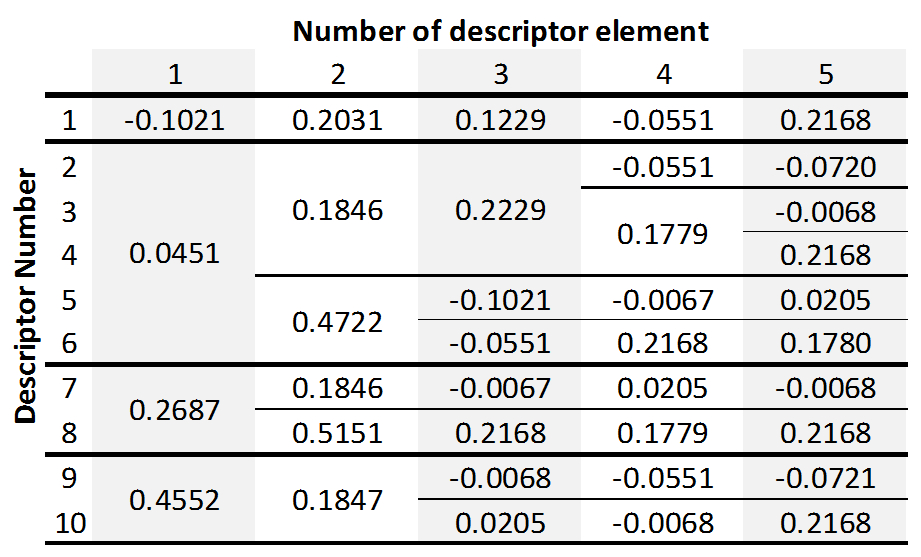}
\caption{Exemplary part of the descriptor dictionary }
\label{fig:Experiment1}
\end{figure}

The elements of descriptors which are similar and their values do not exceed estimated limits, are grouped and will be represented as a single element of the dictionary. An example of grouping is presented in Fig. \ref{fig:Experiment1} for the first element of descriptors with the number between 2 and 6. The rest of descriptor elements, from which another elements are built, are derivatives of the first group. Thanks to grouping, we can decrease the number of similar, duplicated elements of descriptors. 
Thanks to the presented approach, building index of descriptors is also faster, especially when we deal with a very large number of descriptors. The rest of data of keypoints such as position, size or orientation are contained in the last part of the word associated with the descriptor. 
The last step of the process of creation of the dictionary is conversion of data to a binary file as it is sufficient to generate the dictionary only once.

\subsection{Comparison Between Descriptors and Dictionary}
\begin{figure}
\centering
\includegraphics[width=9cm]{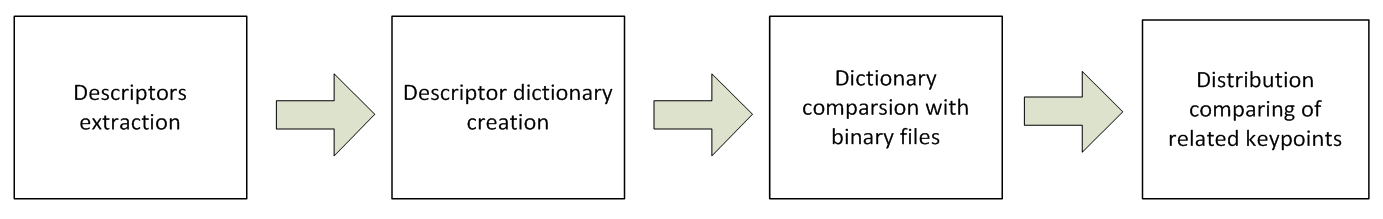}
\caption{Flowchart of image retrieval searching in the set of  images}
\label{fig:Flowchart}
\end{figure}

Every image from the analyzed set has its own descriptor dictionary stored in a form of a binary file (see Section 3.1). Now, let us assume that we have a new query image and we want to find similar images in the large collection of images. The first step is to create a dictionary of its feature descriptors and store it in a binary file. Fig. \ref{fig:Flowchart} presents a flowchart of such image retrieval.  

\begin{figure}
\centering
\includegraphics[width=7cm]{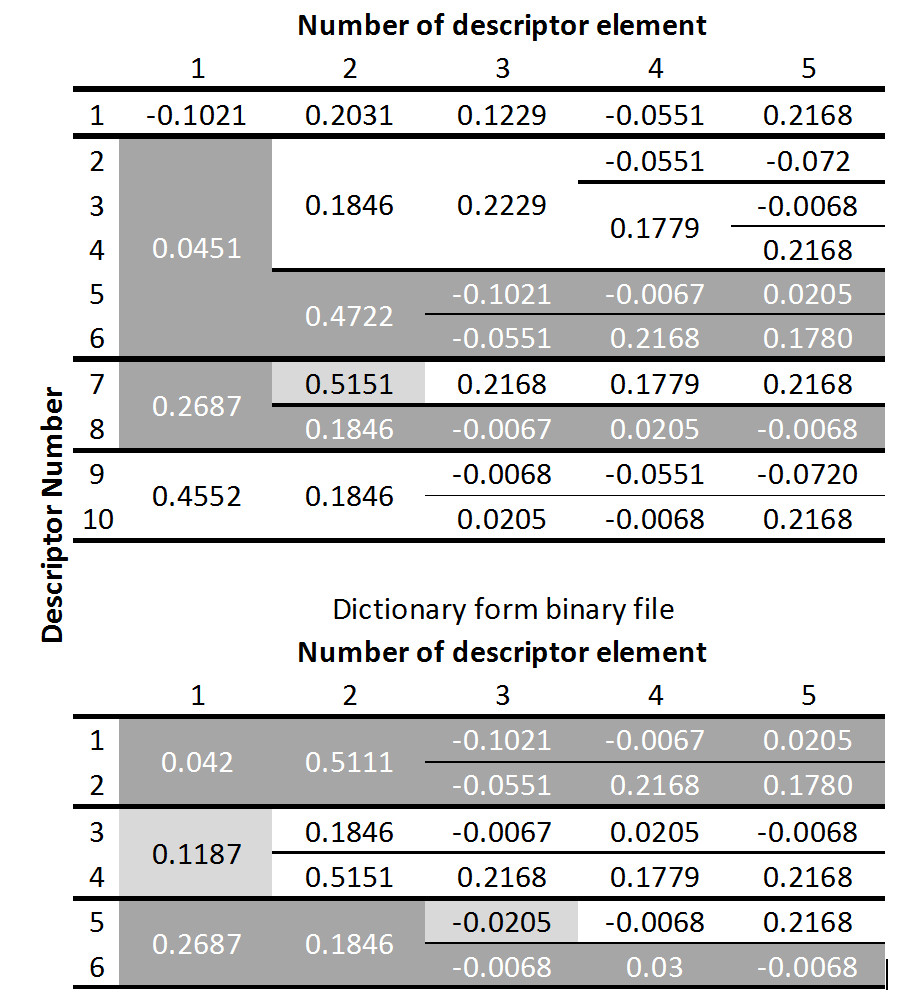}
\caption{Exemplary part of two compared dictionaries}
\label{fig:Samples}
\end{figure}

The next step is a comparison of the query image dictionary with the dictionaries from the binary files. Descriptors values are similar if their sum of absolute differences (SAD) is less than the threshold.
Comparison of two dictionaries is presented in the Fig. \ref{fig:Samples}, where the dark background represents a common part. 

\subsection{Matching Determined Sets of Keypoints}
\begin{figure}
\centering
\includegraphics[width=7cm]{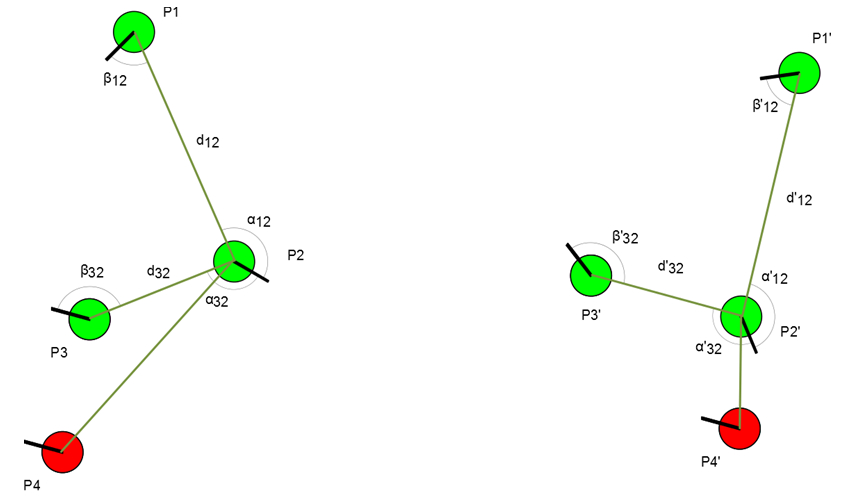}
\caption{Example of keypoints pair checking by mapping them between images.}
\label{fig:Samples2}
\end{figure}

The dictionary comparing process returns a set of pairs of similar keypoints. The next step is to examine keypoint distribution between images. Each pair will be excluded, if their distribution in the relation to the rest of pairs indicates wrong connection. Fig. \ref{fig:Samples2} describes an example of keypoint distribution between two images. 
Each point has its own counterpart in the second set. The method compares the direction and the distance between keypoints from the same set. For example, angles $\beta_{12}$ and $\alpha_{12}$ have the same value as $\beta_{12}$' and $\alpha_{12}$' from the second set. Distances d12 and d12' are also similar. Thus, in this case we can assume, that points P1 and P2 are related. In other case we mark points as not related, e.g. P4 and P4'.

\section{Experimental Results}
\begin{figure}
\centering
\includegraphics[width=7cm]{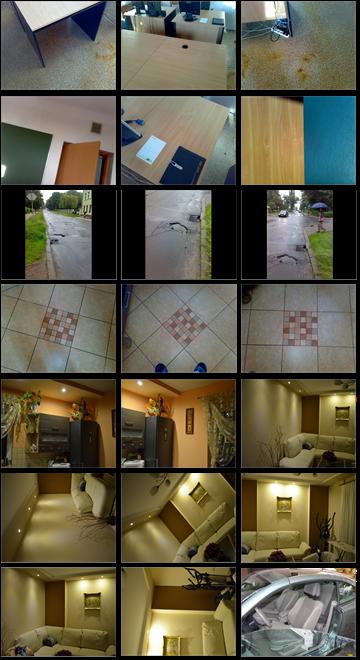}
\caption{Images used in the experiments}
\label{fig:ExperimentalResult}
\end{figure}

In this section we show some examples of the proposed method for content-based image retrieval on the test images presented in Fig. \ref{fig:ExperimentalResult}. For better presentation, we chose images, which are only slightly different. 

Table \ref{Table2} contains test results of comparison between each image with all other from Fig. 6. 
"No. of points" column is the number of descriptors extracted from the image. "Matched" column is the number of related descriptors between current and all other images. "Comparisons" is the number of descriptors compared by using the dictionary. "Combination" is the number of all possible combinations of descriptors between images. As we can see, the number of comparisons  in the proposed method is significantly smaller in relation to the number of all combinations. In our tests, the number of compared descriptors is only 0.18\% of all possible combination.   

\begin{table}
\centering
\caption{Results of comparisons between dictionaries.}
\begin{tabular}{     c     r     r     r     r     c     }
\hline{\smallskip} 
Images & No. of points & Matched & Comparisons & Combinations & Performance \\
\noalign{\smallskip}
\hline
\noalign{\smallskip}
01 & 367 & 225 &19522 &3733124 & 0.52\% \\
02 & 257 & 26 & 3691 & 2614204 & 0.14\% \\
03 & 727 & 103 & 15373 & 7395044 & 0.21\% \\
04 & 80 & 101 & 1747 & 813760 & 0.21\% \\
05 & 408 & 112 & 10773 & 4150176 & 0.26\% \\
06 & 24 & 22 & 413 & 244128 & 0.17\% \\
07 & 729 & 0 & 0 & 7415388 & 0.00\% \\
08 & 414 & 20 & 7676 & 4211208 & 0.18\% \\
09 & 845 & 20 & 7674 & 8595340 & 0.09\% \\
10 & 359 & 128 & 5137 & 3651748 & 0.14\% \\
11 & 318 & 128 & 5107 & 3234696 & 0.16\% \\
12 & 213 & 44 & 3815 & 2166636 & 0.18\% \\
13 & 891 & 52 & 13049 & 9063252 & 0.14\% \\
14 & 785 & 61 & 19567 & 7985020 & 0.25\% \\
15 & 435 & 162 & 10068 & 4424820 & 0.23\% \\
16 & 295 & 95 & 10575 & 3000740 & 0.35\% \\
17 & 489 & 154 & 10408 & 4974108 & 0.21\% \\
18 & 650 & 116 & 14754 & 6611800 & 0.22\% \\
19 & 417 & 186 & 13569 & 4241724 & 0.32\% \\
20 & 464 & 104 & 13479 & 4719808 & 0.29\% \\
21 & 1005 & 5 & 134 & 10222860 & 0.00\% \\
\hline
\end{tabular}
\label{Table2}
\end{table}

\begin{figure}
\centering
\includegraphics[width=5.5cm]{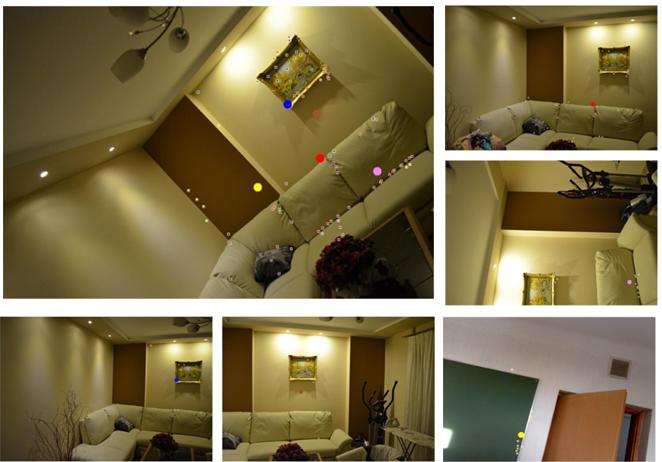}
\caption{Sample of detected groups of descriptors between images}
\label{fig:ExperimentalResult2}
\end{figure}

Fig. \ref{fig:ExperimentalResult2} presents the results of search for common set of keypoints from image number 17. The largest image is our query image. The others are found similar images. Related points are marked on each image. Larger points are the centers of keypoints that describe common area. 

Table \ref{Table3} presents detailed values from comparison procedure between images from Fig. \ref{fig:ExperimentalResult2}. Only part of keypoints were connected, but this number allows selecting a common part of both images. In this case, a single image has been incorrectly marked as related to the query image. It was caused by a similarity between descriptors and their orientation.
\begin{table}
\centering
\caption{Results of comparison between images from Fig.\ref{fig:ExperimentalResult2}}
\begin{tabular}{     c      c      c      c      c      c      }
\hline{\smallskip} 
No. image & No. of  & No. image & No. of  & Math & Comparisons \\
(query) & keypoints & (compared) & keypoints & ~ & ~ \\
\noalign{\smallskip}
\hline										
\noalign{\smallskip}
17 & 489 & 4 & 80 & 31 & 290 \\
17 & 489 & 15 & 435 & 17 & 1491 \\
17 & 489 & 18 & 650 & 20 & 2197 \\
17 & 489 & 19 & 417 & 57 & 1708 \\
17 & 489 & 20 & 464 & 23 & 1723 \\
\hline
\end{tabular}
\label{Table3}
\end{table}

\section{Conclusion}

Analysing results of our test, we can say that creation of the dictionary allows to significantly decrease the number of operations, which have to be done in the process of image comparison. 
In our case, the number of operation has been reduced to the 0.18 of all operations. The approach obtains better results in the case of larger sets of images. Images related to the query image can be found much faster in comparison to the standard all-to-all approach. 
Moreover, saving the dictionary in a binary file allows for more efficient image multiple comparison and reuse of data.

\section*{Acknowledgements}
This work was supported by the Polish National Science Centre (NCN)  within project number DEC-2011/01/D/ST6/06957.

\bibliographystyle{splncs_srt}
\bibliography{biblio}

\end{document}